**Giacomo De Colle, Helena Blackmore, Chris Partridge**

# Constraining ontology mappings using metaphysical choices

## 1. The stratification journey

Cross-ontology mapping is not, usually, a simple lexical matching exercise. It requires an understanding of where the ontologies converge, where they merely approximate one another, and where they differ. Where the ontologies to be mapped both have a top-level foundation, then it is useful to understand where the foundational ontologies converge, where they merely approximate one another, and where they differ. In this paper we discuss the foundations behind a novel methodology for the validation of semantic mappings between different data sources based upon different foundation ontologies, where the methodology builds a framework based upon the metaphysical commitments of the ontologies.

We see mappings as translation mechanisms between two datasets or two data models, with the aim of expressing the semantic content of one source using the model (terms) of the other. Mappings of this kind base themselves on the ontological commitments present in the two data models. For example, one data model might be committed to the existence of books and only refer to books, while the other data model only explicitly commits itself to pages. If we add an explicit mapping postulating that books are collections of pages, then the statement about books of the first model can be expressed using terms from the second model. The mapping can both be expressed at a model level (books are collections of pages) and at an instance or data level (book x in model A is to be rendered as a collection of pages in model B). The ontological commitments of the two models are then the guiding source for the creation of mappings, which allows us to interpret one theory using resources from the other (for more details, see [1][2]).

In many cases, mappings are drawn between data schemas which are not making their ontological commitments explicit. If this is the case, then the person working on the mappings will often have to discover and make these commitments explicit. On the other hand, when one is using higher-level, explicit foundational ontologies as source for data models, the situation is different. A reasonably well-defined ontology is expressing its commitments explicitly. This is an opportunity to facilitate the creation of robust mappings. In this work we do not want to provide a discussion of how to build mappings between ontologies. Rather, we are interested in showing how the metaphysical commitments of two different ontologies or data models can be used in order to constrain and validate a mapping procedure. We will adopt as a use case the mapping between the IES (Information Exchange Standard) [3] and BFO (Basic Formal Ontology) [4] ontologies, and we will focus especially on how the metaphysical choice of unifying and multiplying entities provides cardinality constraints on the mappings between the two. For background, IES is an ontology based on the BORO™ Foundational Ontology [5] and grounded in Extensional 4-Dimensionalism, while BFO combines both three-dimensional and four-dimensional aspects.

In order to introduce our method, we need to first define the terms "unification" and "division". Take the notorious metaphysical case of the statue and the clay. A unifying ontology makes the metaphysical choice of only counting one item in the case of the statue and the clay, without deciding to make any distinction between them. On the other hand, a dividing ontology would make a different metaphysical decision, for example by differentiating between form and matter, and include two items in their ontology.

The method we adopt in this work takes the form of a *stratification journey*. The intuition is the following: starting from the most metaphysically unified item possible, we can imagine a journey composed of a series of choices of metaphysical divisions. The more one follows the path of the journey, the more division choices they make, the more their ontology will become stratified and include multiple objects. On the other hand, the ontologist who decides to stop at a higher level in the decision tree will find themselves with a more unified ontology. The decision tree then allows us to characterize different ontologies on the basis of metaphysical choices which have the consequence of unifying or multiplying entities. The stratification journey graph, depicted in Figure 1, is then not itself representing an ontology or a hierarchy, but rather a classification mechanism. This strategy is a direct development of the work discussed in [6].

The stratification journey begins with the most unified ontological entity: the supersubstantival object [7]. A supersubstantival object is the fusion of a portion of matter *and* spacetime, and does not distinguish between the two of them. In the metaphysical picture drawn at this level, there is no distinction between me, the temporal span of my life and the portion of space which I occupy during my life. Starting from this situation, there are a series of choices which one can make in dividing their ontology. The first stratification we decided to include in the graph is the division between temporal and non-temporal entities. These entities have in different occasions been called endurants and perdurants [8], or continuants and occurrents [4]. Each one of them can then be divided by stratifying the object of interest and its container. For example, on the endurant side, we can divide between space and the material entities which occupy a position in it. And on the perdurant side, we can divide between (space)time and the processes and events which occupy a position in it.

Continuing down the endurant path, the space-occupying entities can furthermore be divided into material entities (chairs, people, animals, and so on) and *sites*, i.e. Aristotelian places [4]. Continuing down the perdurant path, we can divide items occupying spacetime following a mereological criterion: following BFO [4] we can call the time-occupying items with no temporal parts, i.e. instantaneous events, *process boundaries*. The other time-occupying occurrents are called processes. On the other hand, the containing items can be divided between time and spacetime. This division process is by no means exhaustive of the entire possibilities of metaphysical division. Mostly, this diagram touches upon stratifications made on the basis of principles related to spacetime, location and mereology. Nevertheless, it can already be fruitfully employed to showcase the principles behind it and to be applied in the case of mappings between BFO and IES.

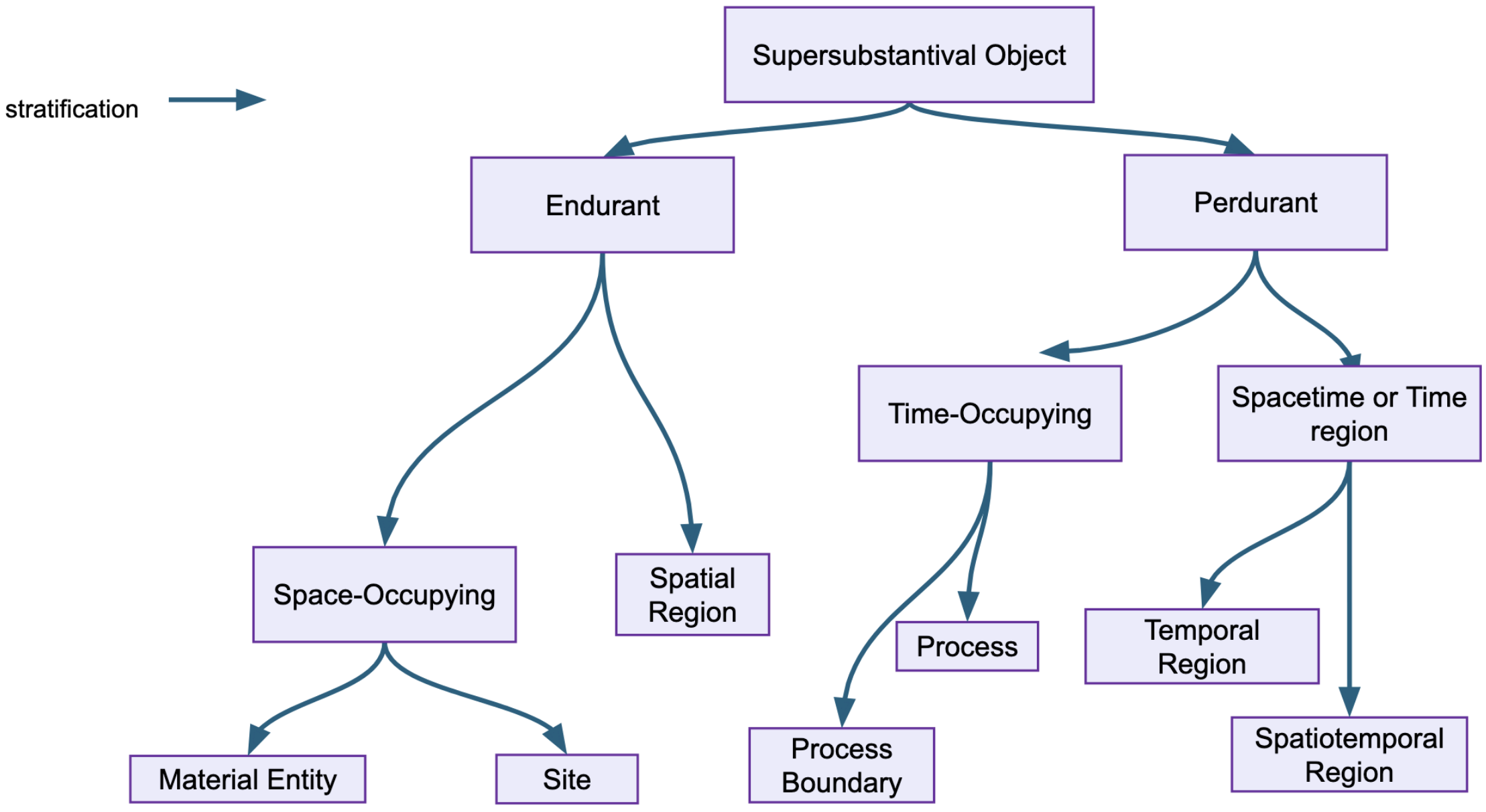

**Figure 1** The stratification journey.

After designing the stratification journey, it is relatively easy to identify relations of ontological dependence between the stratified entities. For example, it is impossible for a material entity to exist without it also having a location in space. The supersubstantivalist believes that spacetime is the fundamental and perhaps only substance. Yet when the other ontologists make a metaphysical division on the supersubstantival object, they make a division which in principle should not produce different modal profiles of the divided entities, in the sense that the process of someone's life is intimately connected to the endurant object that that person is, so that there is a type of ontological dependence between the two. When one of them is present, the other also is. In Figure 2 we show a representation of the ontological dependence between the objects divided in metaphysical splits.

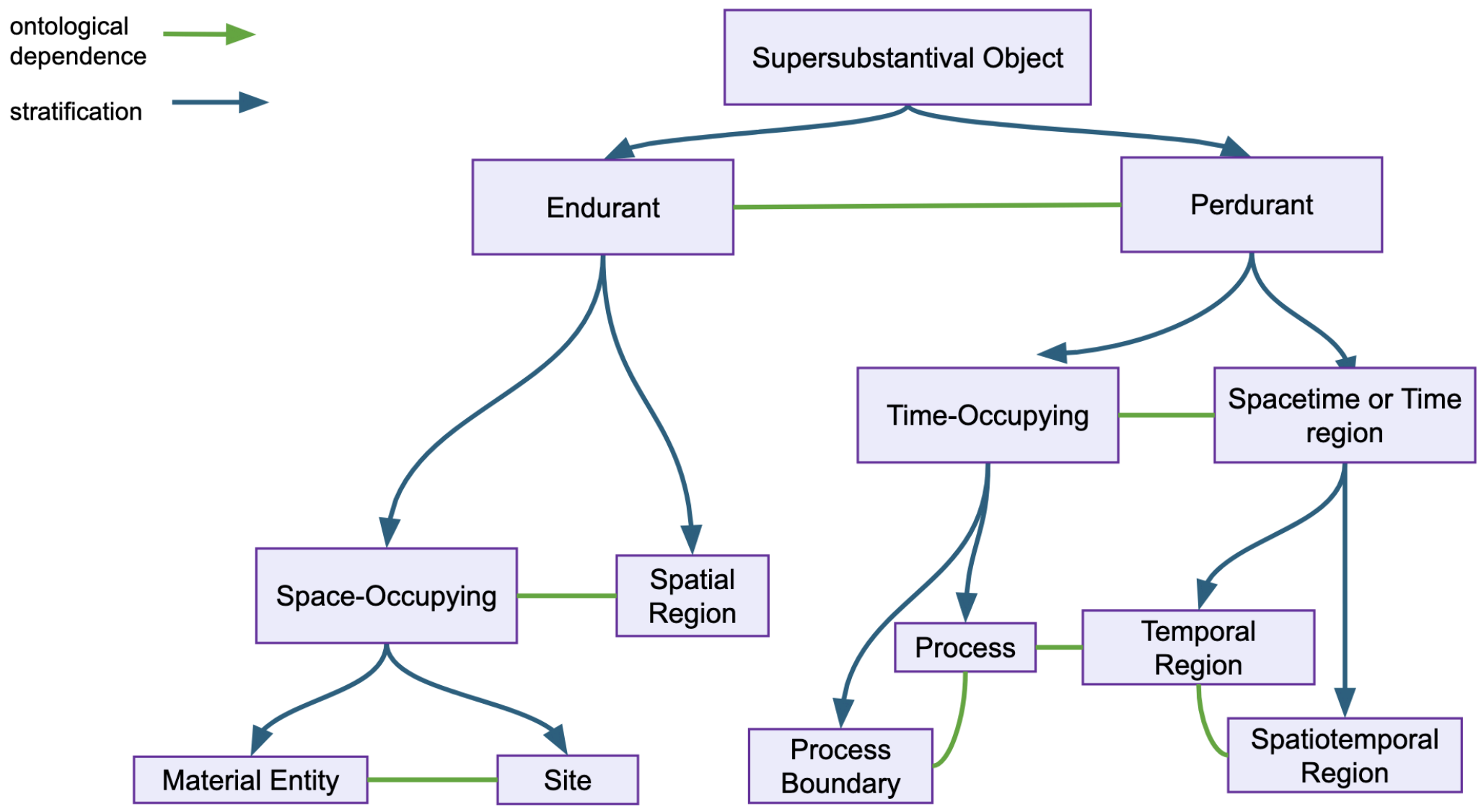

**Figure 2** Dependency relations between stratified objects.

Once this setup is given, a first obvious consequence can be identified. If a supersubstantivalist ontologist A represents an object *x*, the ontologist B who has decided to traverse the first level of the graph will now have to decide whether to create an endurant *y* or a perdurant *z* in order to create a representation which is semantically equivalent to the first. Moreover, given the nature of ontological dependence, they will in any case have to create yet another object. If a third ontologist decides to map the model created with one ontology into the model created with the other, they will have to respect a cardinality constraint: if they are mapping from A to B, they will have to multiply objects, as one object in A requires two objects in B. If they are mapping from B to A, they will have to reduce objects, given that two objects in B are equivalent to one object in A. In BFO and IES terms, for example, let's take the case where we are mapping a car in IES to BFO. In IES, the car is a physical, spatiotemporally extended entity. In BFO, the car will produce at least two entities: the physical car, which is only extended in space, and its spatiotemporally extended history. The metaphysical choices made by the two ontologists then create cardinality constraints on mapping procedures which can be used to validate mappings between the two ontologies.

Another principle follows from the ones we have already discussed. Imagine that ontologist A from the previous paragraph creates two supersubstantival objects and a relation of parthood between them in their ontology. Given what we have already discussed in the previous paragraph, the two items become at least four items in ontology B. But if this is the case, then the relations between them need to correspondingly multiply, according to a principle of mereological harmony [9]. Cardinality rules then not only apply to items in one's ontology, but also to relations between them. Together, rules on the multiplication of entities and of multiplication of relations provide a useful framework. Our hypothesis is that these metaphysical constraints can be fruitfully adopted in a computational setting to facilitate data transformation. In the next section, we show a practical implementation of this principle.

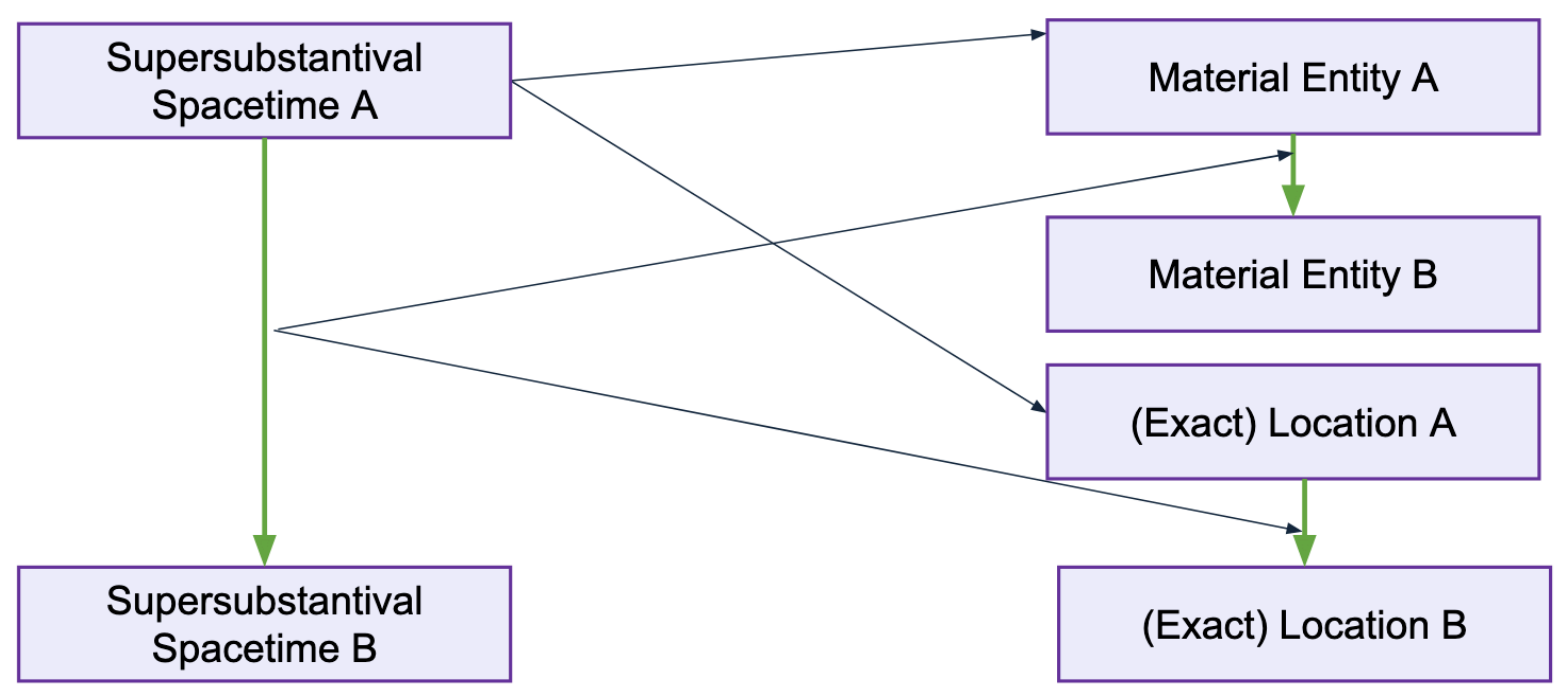

**Figure 3** The impact of the stratification journey on the mapping (black arrows) of relations (green arrows).

## 2. Validating mappings with cardinality constraints

The use case we decided to adopt in order to test our hypothesis is the one of a mapping between the IES and BFO ontologies. The setting is simple: IES adopts an explicit supersubstantivalist, unifying approach. BFO, on the other hand, follows the stratification journey entirely and divides entities adopting both an endurantist and a perdurantist perspective or, in BFO terms, distinguishes between continuants and occurrents. The differences and similarities between the two have been discussed in [10] and an ongoing effort between the two communities exists in order to create robust semantic mappings at a class level. Figure 4 shows a slightly modified version of the stratification journey, adapted to show how BFO and IES classes can be characterized on the basis of their metaphysical choices.

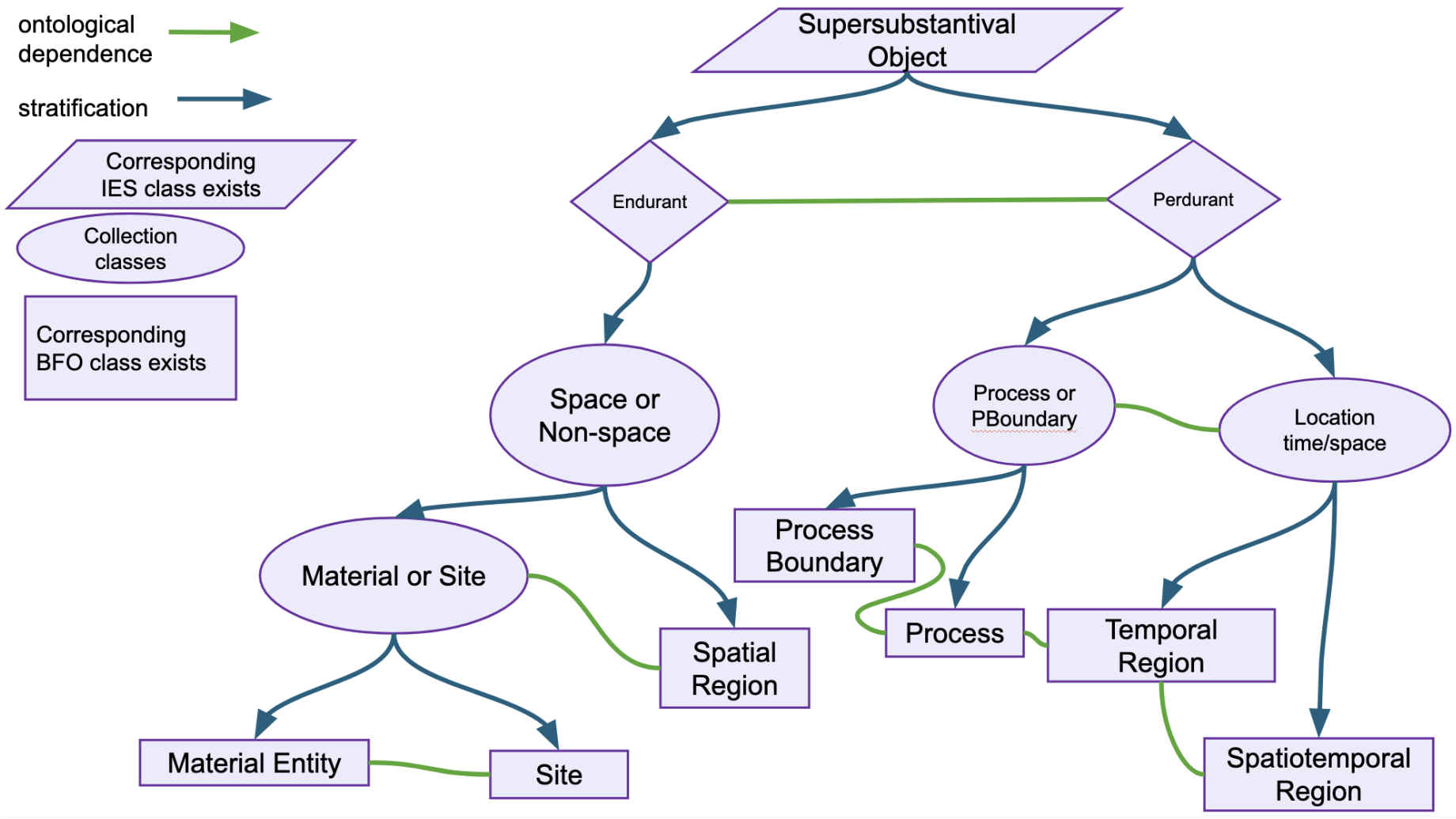

**Figure 4** Applying the stratification journey to IES and BFO.

In the rest of the section we extend this work to an example-driven approach, where we try not to map IES and BFO classes, but rather to transform instance-level data from one ontology into the other. The example we decided to start from is Winston Churchill's birth. In IES, he is represented as a supersubstantival entity, that is, a part of spacetime extended in four dimensions and including matter. In BFO, the usual process of entity stratification takes place. In Figure 5 we show how this approach is applied to the specific use case with one difference from the previous diagrams. While theoretically speaking BFO does divide entities compared to IES, for practical purposes it is often helpful to concentrate only on a selected number of instances that are important to generate in BFO space, which we mark in green in the graph.

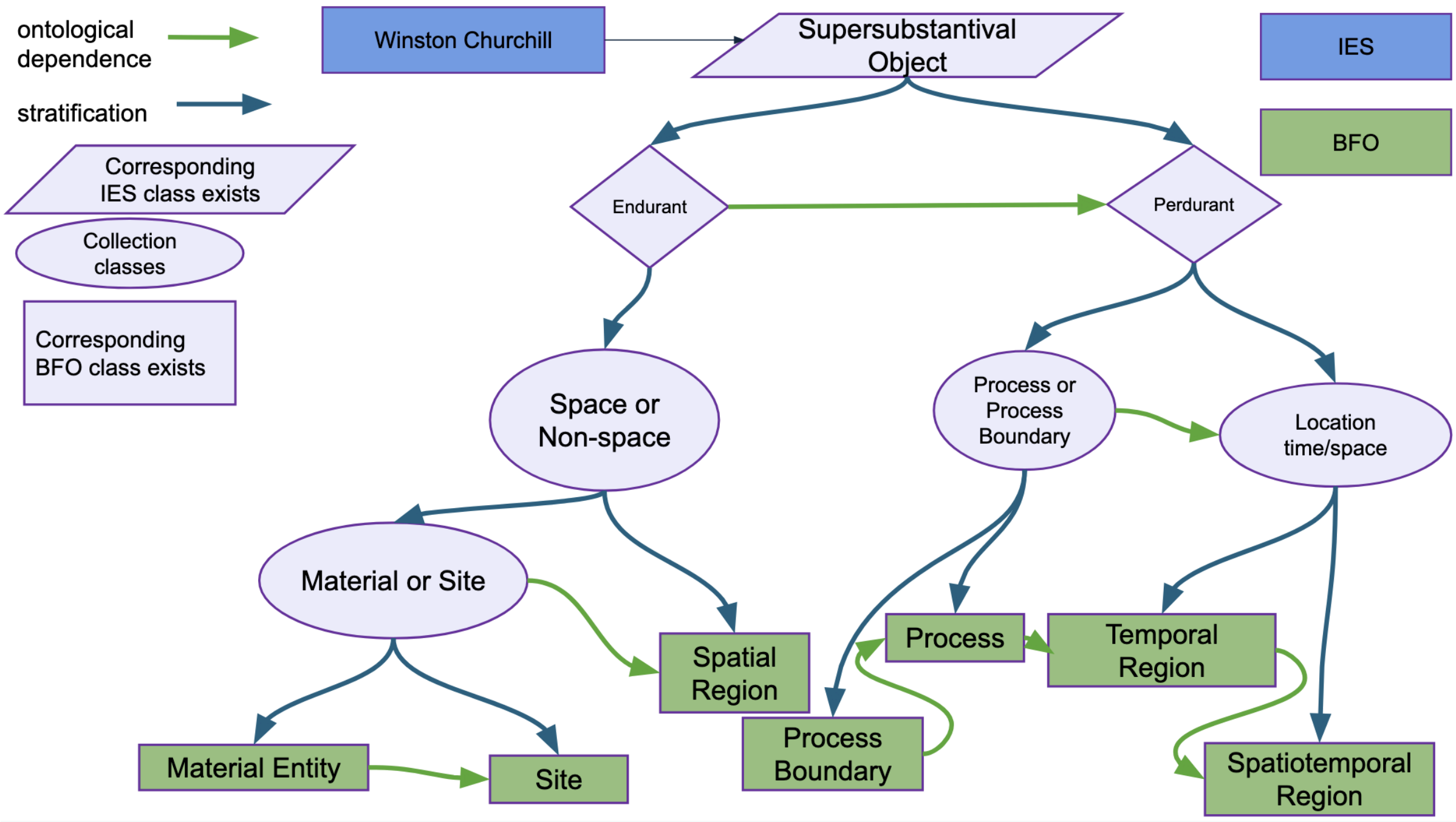

**Figure 5** Stratifying a single instance, pragmatically choosing what are the most important entities to generate through the mapping.

Given this setup, we proceeded to test the feasibility of this approach computationally. We first created small IES and BFO RDF graphs to encode examples such as the one described in Figure 5. The examples were deliberately kept small and focused, but captured several key classes and patterns in both ontologies. We then used an open-source python library for transforming RDF data from one form to another; in most cases: data from one ontology to another. Configuration of this transformation is provided as a series of SPARQL queries [11], and was used to generate IES knowledge graphs from BFO graphs and vice versa. This allowed us to examine and test the mapping process in both directions and to assess whether the information in both graphs was preserved and consistent. After the mapping and translation, we annotated the generated RDF triples with information about their origin in the other ontology in order to easily mark what their class of origin was. Separately, we wrote a simple Python script that includes a SPARQL query encoding the cardinality constraints we previously discussed. For example, an IES:object needs to be expanded to a minimum of 2 BFO objects (the object and its history). The workflow is presented in Figure 6. More ad-hoc pragmatically derived constraints can be specified depending on the situation. For example, following Figure 5, Winston Churchill's birth needs to generate at least five BFO entities: the material entity of Winston Churchill, the site and spatial regions he occupies at his birth, the process of his birth, the instance at which this process starts, and the temporal and spatiotemporal regions occupied by this process. We then ran the SPARQL queries on the results of the mapping, flagging the presence of triples that did not follow the cardinality constraints.

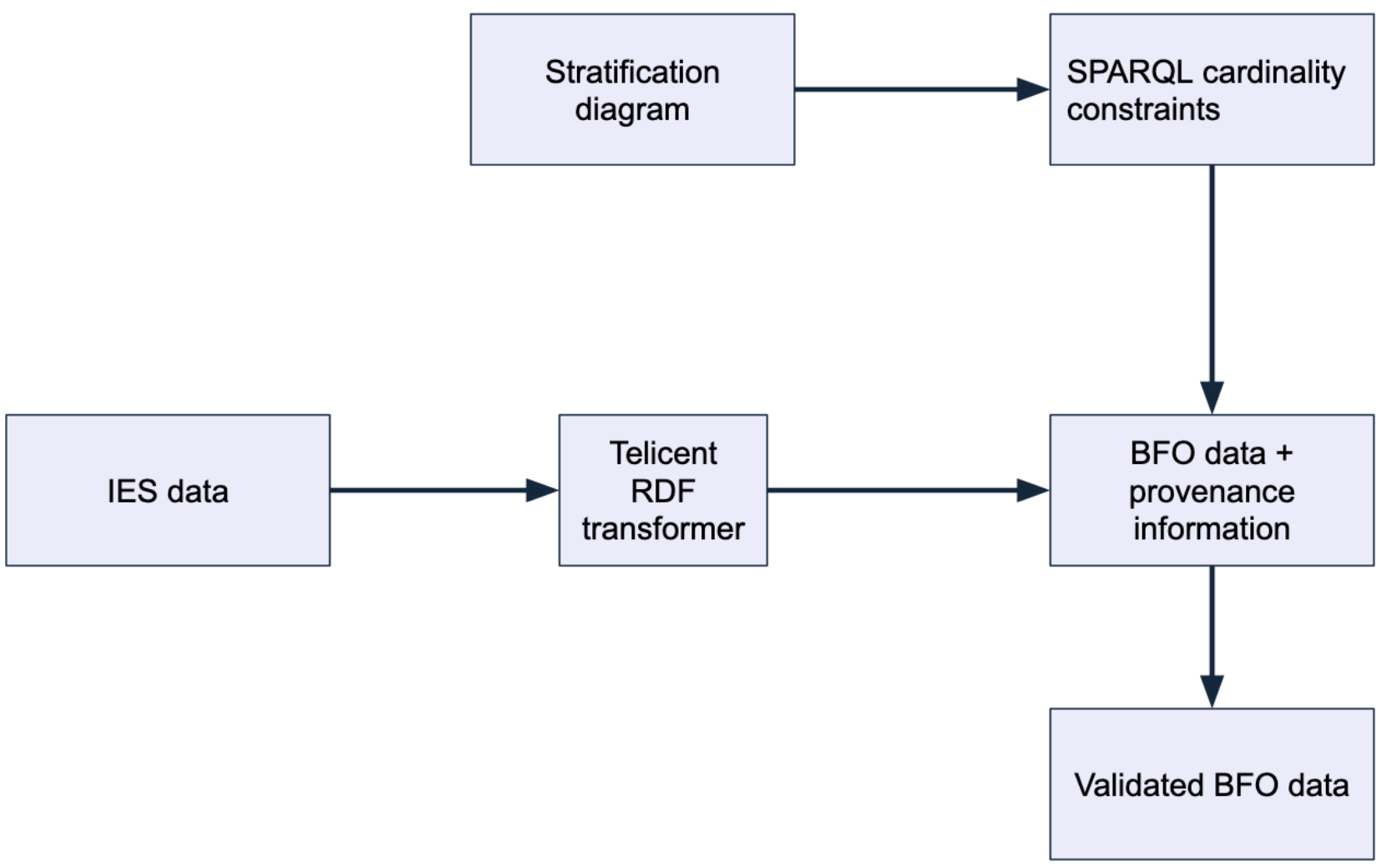

**Figure 6** The workflow of the implemented test case.

To conclude, we have argued that mappings between datasets are based on the metaphysical commitments of the ontologies that regulate them, and that these commitments can be used in order to constrain and validate such mappings. In cases where the ontologies are explicit about their metaphysical commitments, it is general principles coming from metaphysics that can guide the creation of such constraints. In this paper we focused on the process of ontological stratification, understood as a method to understand how the multiplication or unification of entities occurs in different ontologies. Furthermore, we discussed as an example the test case of mappings between IES and BFO, and we especially focused on providing cardinality constraints on the mappings between the two ontologies. In order to demonstrate the applicability of our method, we showcased how these principles can be operationalized through SPARQL queries validating the results of a mapping pipeline.